\documentclass[runningheads]{llncs}
\usepackage{graphicx}
\usepackage[utf8]{inputenc}
\usepackage{epsfig}
\usepackage{calc}
\usepackage{amssymb}
\usepackage{amstext}
\usepackage{amsmath}
\usepackage{multicol}
\usepackage{multirow}
\usepackage{pslatex}
\usepackage{amsmath}
\usepackage{listings}
\usepackage[ruled,vlined,linesnumbered]{algorithm2e}
\usepackage{comment}
\usepackage{amssymb}
\usepackage{url}
\usepackage{enumitem}
\usepackage{hyperref}

\newcommand{\mapfr}{MAPF\ensuremath{^{R}}\xspace}

\begin{document}

\title{Plan Execution for Multi-Agent Path Finding with Indoor Quadcopters}
\titlerunning{Plan Execution for MAPF with Indoor Quadcopters}


%
%
\author{Matou\v{s} Kulhan and Pavel Surynek\orcidID{0000-0001-7200-0542}}
%
\authorrunning{M. Kulhan and P. Surynek}

%
\institute{
Faculty of Information Technology\\
Czech Technical University in Prague\\
Th\'{a}kurova 9, 160 00 Praha 6, Czechia\\
\email{\{kulhama7, pavel.surynek\}@fit.cvut.cz}
}

\institute{
\phantom{.}
\email{\phantom{.}}
}

\maketitle

\begin{abstract}
We study the planning and acting phase for the problem of multi-agent path finding (MAPF) in this paper. MAPF is a problem of navigating agents from their start positions to specified individual goal positions so that agents do not collide with each other. Specifically we focus on executing MAPF plans with a group of Crazyflies, small indoor quadcopters . We show how to modify the existing continuous time conflict-based search algorithm (CCBS) to produce plans that are suitable for execution with the quadcopters. The acting phase uses the the Loco positioning system to check if the plan is executed correctly. Our finding is that the CCBS algorithm allows for extensions that can produce safe plans for quadcopters, namely cylindrical protection zone around each quadcopter can be introduced at the planning level.

\keywords{path finding, planning, acting, multiple agents, indoor quadcopters, Crazyflie, MAPF, localization}
\end{abstract}

\section{Introduction}

\begin{figure}[h]
    \centering
    \includegraphics[trim={0.5cm 22cm 8.8cm 2.3cm},clip,width=0.8\textwidth]{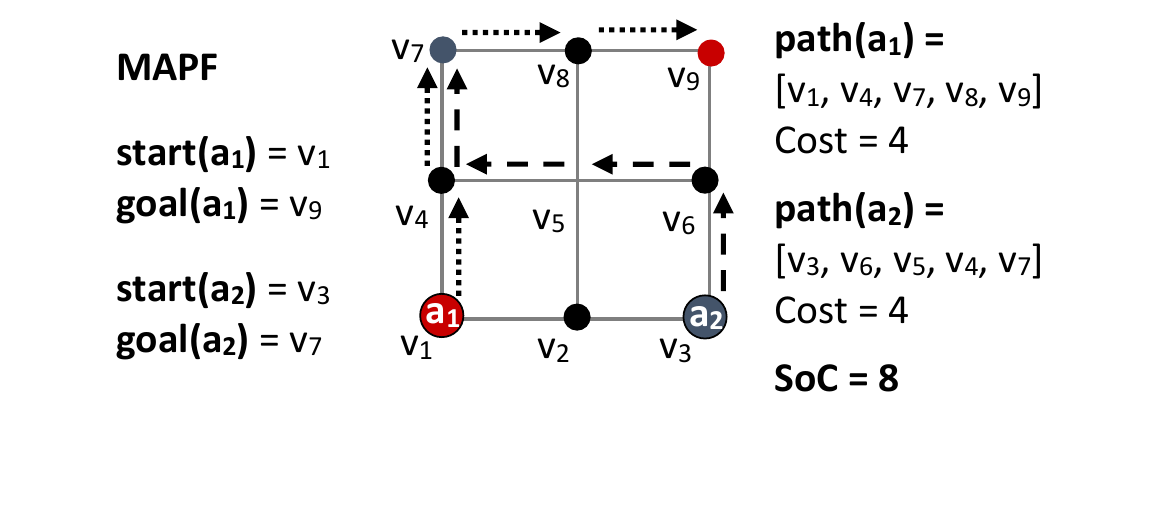}
    \vspace{-1.0cm}\caption{A MAPF instance with two agents $a_1$ and $a_2$.}
    \label{fig:MAPF}
\end{figure}

In {\em multi-agent path finding} (MAPF) \cite{DBLP:conf/focs/KornhauserMS84,DBLP:journals/jair/Ryan08,DBLP:conf/icra/Surynek09,DBLP:journals/jair/WangB11,SharonSFS15,DBLP:conf/socs/FelnerSSBGSSWS17} the task is to navigate agents $A=\{a_1,a_2,...,a_k\}$ from their starting positions to given individual goal positions so that they do not collide with each other. The standard discrete version of MAPF takes place in an undirected graph $G=(V,E)$ whose vertices represent positions and edges define the topology of the environment - agents move across edges, but no two agents can reside in the same vertex at a moment, nor two agents can traverse an edge in opposite directions (Figure \ref{fig:MAPF}). There are numerous applications of the problem ranging from warehouse logistics to game development \cite{DBLP:journals/aimatters/MaK17}.

Recent progress in MAPF brings the abstract problem closer to real life applications. Concretely a variant of MAPF that integrates continuous aspects of the real world has been devised - MAPF with continuous time (\mapfr) where agents still move between the finite number of vertices but the vertices are embedded in a metric space (continuous 2D or 3D space) and the movements are no longer discrete. That is, agents move smoothly between the positions in the metric space usually along straight lines instead of instantaneous skips between the neighboring vertices as in MAPF. Agents can be of any shape in \mapfr and the collision between agents is defined as any overlap between their bodies. Collisions are avoided in the time domain by allowing an agent to wait at a position for certain amount of time.

\begin{figure}[!ht]
	\centering
	\begin{minipage}{0.49\textwidth}
		\includegraphics[width=\textwidth]{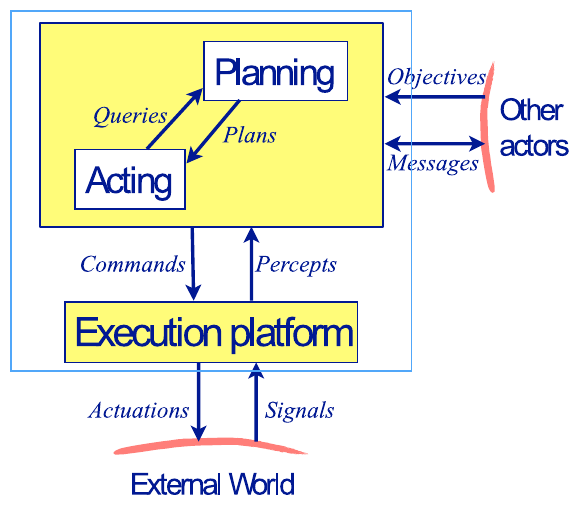}
		\caption{A general scheme of planning and acting in intelligent agents \cite{planning_acting}.}
		\label{img:actor}
	\end{minipage}
	\begin{minipage}{0.49\textwidth}
		\includegraphics[width=\textwidth]{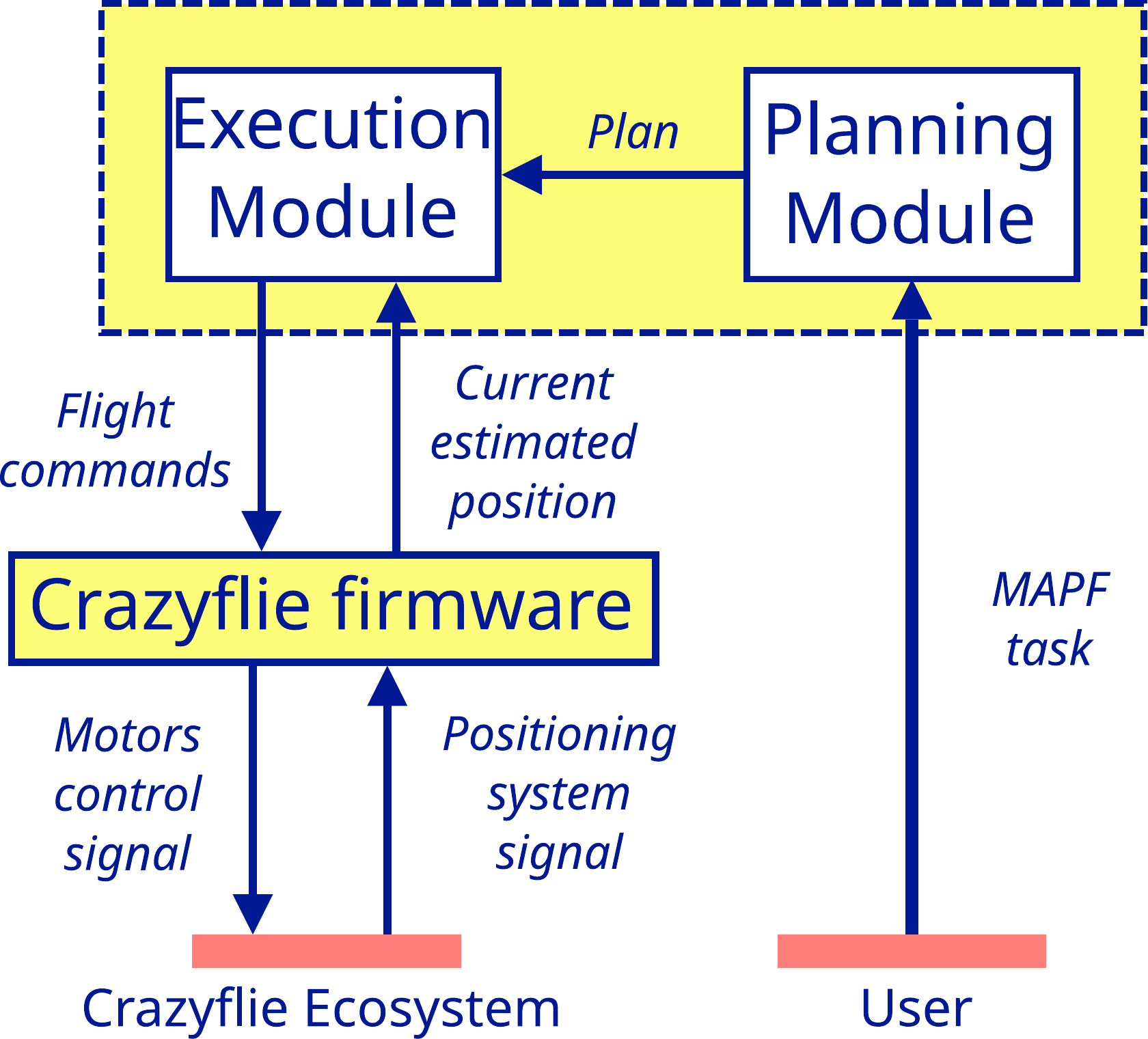}
		\caption{Planning and acting for Crazyflie quadcopters.}
		\label{img:actor_spec}
	\end{minipage}
\end{figure}

We implement the {\bf planning-acting} scheme from \cite{planning_acting} (Figures \ref{img:actor} and \ref{img:actor_spec}) and demonstrate that planning algorithms for \mapfr are suitable for constructing plans that are executable by real {\em robotic agents} - Crazyflies, small indoor quadcopters coupled with localization system our case. Thus we show that contemporary \mapfr algorithms are ready for transfers into real-life applications.

\section{Related Work and Background}

Previous attempts to bridge the theory and acting with real robots for MAPF include the use of mobile robots Ozobot EVO \cite{DBLP:conf/aaai/ChudyS21}. The planning phase was represented by the standard discrete MAPF. Hence the output plan had to be post-processed to form continuous command sequences before it was executed by the robots. The disadvantage of this approach is that plans are searched in a discrete space that is different from the continuous target environment which may not keep the desired properties of the plan such as optimality.

In this work, we use \mapfr model that allows for planning and acting in identical spaces - in a grid embedded in a continuous 3D space in our case. Existing planning algorithm for \mapfr, namely Continuous-time Conflict Based Search (CCBS) \cite{DBLP:journals/ai/AndreychukYSAS22} extends the previous
Conflict Based Search (CBS) \cite{SharonSFS15} by resolving conflicts between agents in the time domain. Original CBS uses lazy resolution. It first plans one path for each of the agents which was found by single-agent path-finding algorithm ignoring other agents. If a collision is detected in the plan, say between two agents $a_i$ and $a_j$ in vertex $v$ at time $t$, then the search branches with two new constrains one for each branch. The first prohibits agent $a_i$ from occupying $v$ at $t$ and second one agent $a_j$. The search in each branch continues by single-agent path-finding which now must satisfy the constraints.

CCBS follows the same framework as CBS with several differences. CCBS constraints are imposed over move actions and time intervals instead of vertex and time pairs. Instead of discrete path-finding algorithms, CCBS uses the SIPP \cite{DBLP:conf/icra/PhillipsL11} algorithm for single-agent path-planning that plans w.r.t. {\em safe time intervals} assigned actions and allows an agent to wait in a vertex to avoid executing a move action within an unsafe interval. Unsafe intervals are calculated geometrically from the shape of agents.

\section{Planning and Acting for \mapfr with Quadcopters}

We modified the original CCBS implementation \cite{ccbs:implementation} to support 3D grids and added collision detection mechanism for agents moving in 3D.
As the quadcopters usually cannot fly too close to each other and must keep relatively bigger vertical distances we model the agents as tall {\bf cylinders} with their base parallel to the $xy$-plane (the actual quadcopter is assumed to be in the center of the cylinder).

Collision detection of two cylinders is done by splitting the task into a two 2D collisions. First we calculate an unsafe interval in the $xy$-plane. This is done analytically by determining when the two moving circles (with same radius as the cylinders) will collide. Next we calculate an unsafe interval of two line segments (with lengths equal to the heights of the cylinders) in the z coordinate. Intersection of these two intervals then represents an unsafe interval for the cylinders.

\subsection{Acting Hardware: Crazyflie Ecosystem}

\begin{figure}[!ht]
	\centering
	\begin{minipage}{0.39\textwidth}
		\includegraphics[width=\textwidth]{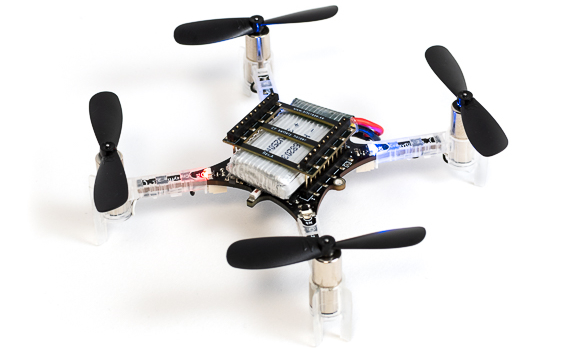}
		\caption{Crazyflie~2.1 \cite{crazyflie:ecosystem}}
		\label{img:crazyflie_2_1}
	\end{minipage}
	\begin{minipage}{0.6\textwidth}
		\includegraphics[width=\textwidth]{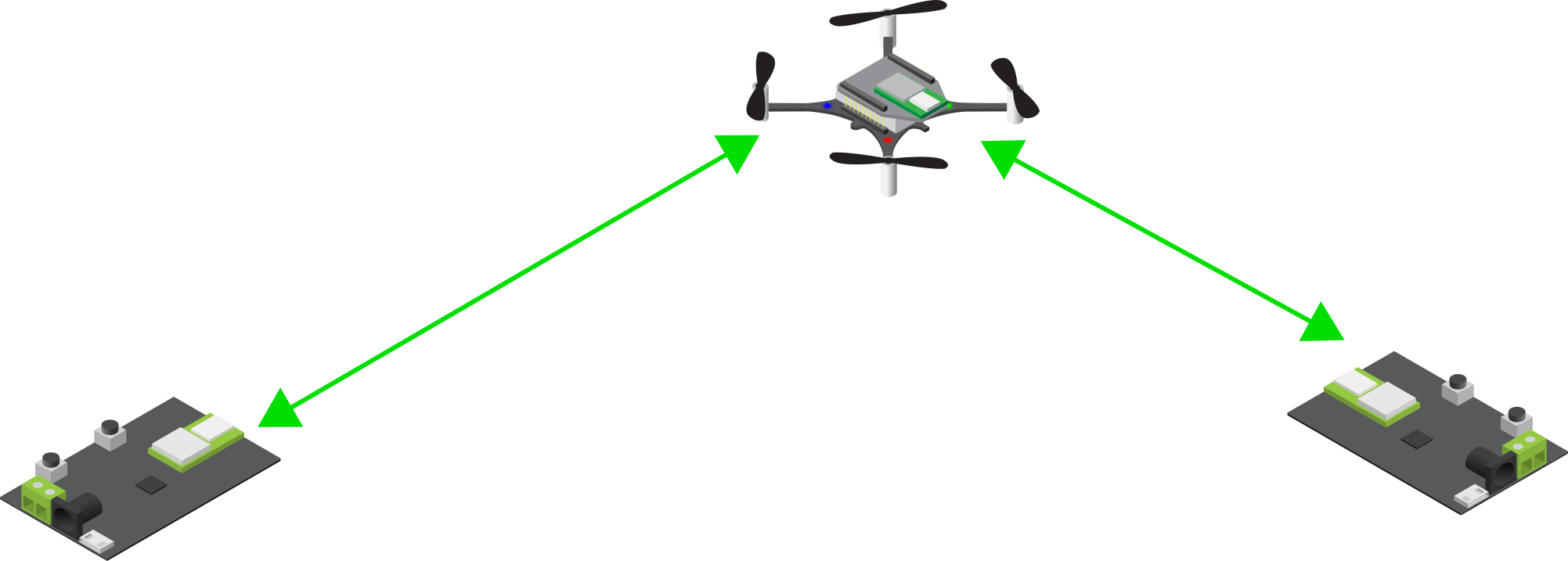}
		\caption{Loco positioning system \cite{crazyflie:ecosystem}}
		\label{img:loco}
	\end{minipage}
\end{figure}

For the demonstration of acting for \mapfr we used Crazyflie 2.1, small indoor quadcopter (Figure \ref{img:crazyflie_2_1}). Crazyflie comes with hardware ecosystem  \cite{crazyflie:ecosystem} that can be divided into three areas: (i) Crazyflie Family, consisting of several versions of Crazyflie quadcopters, (ii) positioning systems, consisting of external sensors to determine position of Crazyflies - in this work we tested the Loco positioning system (see Figure \ref{img:loco}) based on measuring distance to anchors, specified accuracy of 10 cm, and (iii) technologies for remotely controlling the Crazyflies, this includes USB radio dongle Crazyradio PA and {\tt cfplib}, a Python library for sending commands using the radio.

\subsection{Acting Software: Crazyflie Plan Execution Module}

Plan execution module is represented by a Python program using the cflib Python library. The module periodically receives most current estimated position from each controlled Crazyflie and sends them individually generated flight commands based on the desired flight plan. In this paper we propose and compare three different methods for generating these commands: (1) {\bf BHL method} - uses High Level Commander, a firmware module, which receives abstract commands containing absolute position and duration and refines them into setpoints on board of the Crazyflie, (2) {\bf BLL method} - generates the setpoints directly and sequentially send them to the Crazyflie (the pseudo-code of of BLL is shown as Algorithm \ref{alg:BLL}), (3) {\bf VLL method} - checks if the Crazyflie is inside a bounding box around the desired coordinates and if not sends a command to move in the direction of these coordinates.



    
        
    

\begin{algorithm}[t]
\begin{footnotesize}
\SetKwBlock{NRICL}{BLL-Execute($P(a_i)$)}{end} \NRICL{
    $(x_l, y_l, z_l, t_l) \gets$ first-position($P(a_i)$) \\

    \For{each position $(x, y, z, t) \in P(a_i)$} {
       $d \gets t - t_l$;
       $v_x \gets \dfrac{x - x_l}{d}$;
       $v_y \gets \dfrac{y - y_l}{d}$;
       $v_z \gets \dfrac{z - z_l}{d}$\\
       
       \While{current-time() $< t$}{
           $t_r \gets $current-time()$ - t_l$;
           $x_c \gets x_l + v_x \cdot t_r$;
           $y_c \gets y_l + v_y \cdot t_r$;
           $z_c \gets z_l + v_z \cdot t_r$\\
           send-command($i$,$x_c$, $y_c$, $z_c$)\\
           delay()\\
       }
       $x_l \gets x$;
       $y_l \gets y$;
       $z_l \gets z$;
       $t_l \gets t$\\
    }
}
\caption{BLL method for execution of \mapfr plan $P(a_i)$ for agent $a_i$.} \label{alg:BLL}
\end{footnotesize}
\end{algorithm}

\section{Experimental Evaluation}


All experiments were performed in 2 $\times$ 2 $\times$ 2 m flying area with 8 Loco positioning anchors, one in each corner of the area. Position of each Crazyflie was logged every 10 ms. The first experiment focused on testing of the three plan execution methods and evaluating their accuracy w.r.t. flight plan produced by CCBS. The three other experiments tested the plan execution with three different \mapfr tasks with variable number of quadcopters.

\subsection{Results}

\begin{figure}[!ht]
	\centering
	\begin{minipage}{0.39\textwidth}
        \centering
        \begin{tabular}{| c || c | c |}
            \hline
            \multirow{2}{*}{Method} & \multicolumn{2}{c |}{Error} \\ \cline{2-3}
            & Max. & Avg. \\ \hline \hline
            BHL & 0.644 m & 0.223 m \\ \hline
            BLL & 0.662 m & 0.241 m \\ \hline
            VLL & 0.601 m & 0.282 m \\ \hline
        \end{tabular}
        \caption{Results of experiment 1}
        \label{table:results}
	\end{minipage}
	\begin{minipage}{0.59\textwidth}
		\includegraphics[width=\textwidth]{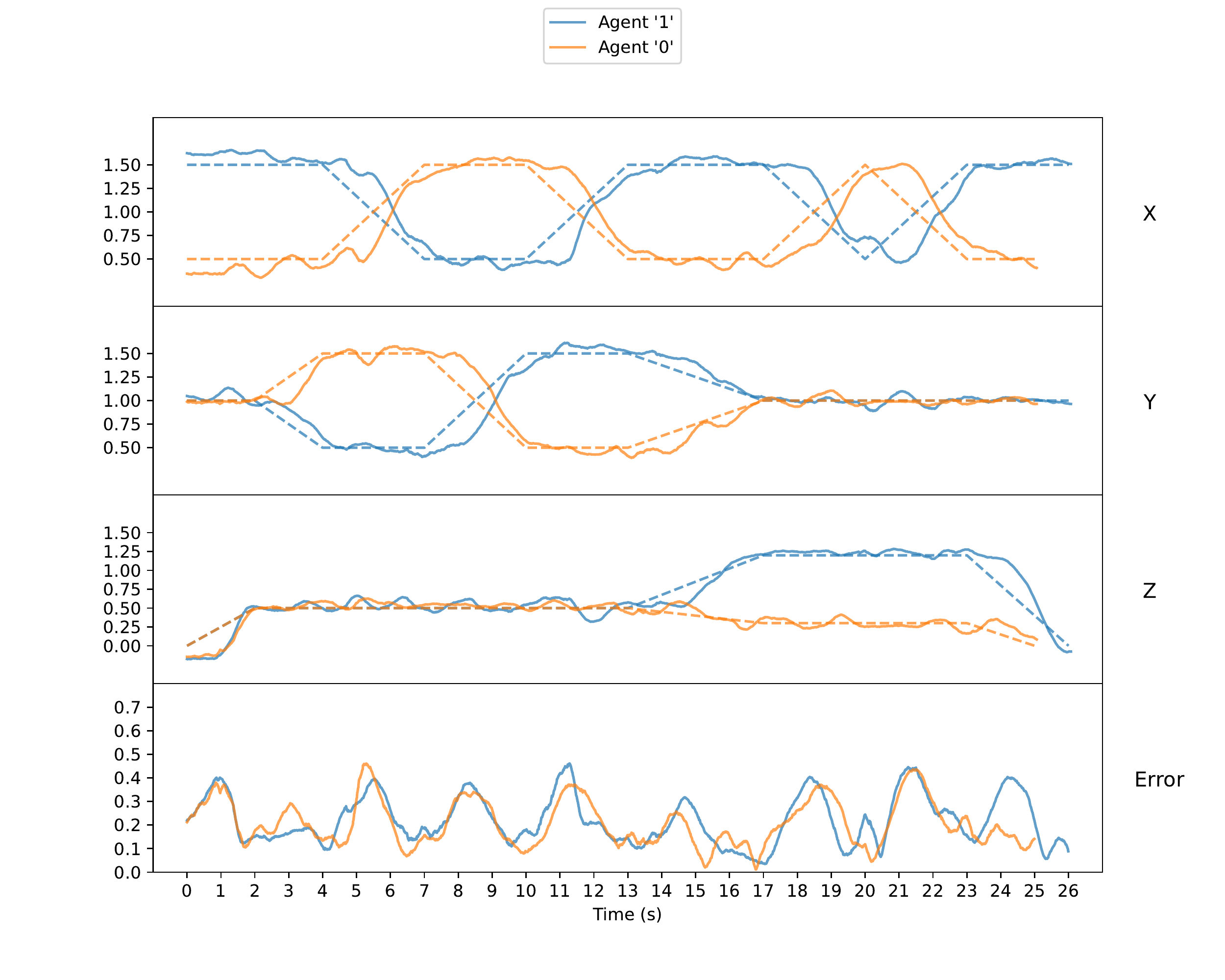}
		\caption{Error over time for Experiment 1}
		\label{img:experiment_plot}
	\end{minipage}
\end{figure}

In the first experiment the flight plan was successfully executed 13 times. Aggregated results are shown in Table \ref{table:results}. For each of the successful execution we plotted error and position of the agents over time. One of these plots can be seen in Figure \ref{img:experiment_plot}. Video recording of all four experiments can be seen on: \href{https://youtu.be/-ftfsM71aY4 }{https://youtu.be/-ftfsM71aY4}. 

Our key finding is that the proposed planning-acting system for \mapfr is capable of generating optimal flight plans and executing them using the Crazyflie Ecosystem with high success rate. We also found that all three proposed methods are close in terms of accuracy with BHL having the lowest average error and VLL the lowest maximum error. 

\section{Conclusion}

We demonstrated that \mapfr planning algorithms require only minor modifications for being successfully applied for plan generation and acting with real robotic agents - in our case small indoor quadcopters Crazyflie. This shows maturity of the \mapfr technology and feasibility of deployments of real-life \mapfr planning-acting systems. For the future work we plan to extend our experiments with more diverse scenarios and localization systems.


\section*{Acknowledgements}
\noindent 
This research has been supported by GA\v{C}R - the Czech Science Foundation, grant registration number 22-31346S. 

\bibliographystyle{splncs04}
\bibliography{references}

\end{document}